# Mining customer product reviews for product development: A summarization process

Tianjun HOU, corresponding author, tianjun.hou@centralesupelec.fr, Laboratoire Genie Industriel, CentraleSupélec, Université Paris-Saclay, Gif-sur-Yvette, France;
Bernard YANNOU, bernard.yannou@centralesupelec.fr, Laboratoire Genie Industriel, CentraleSupélec, Université Paris-Saclay, Gif-sur-Yvette, France;
Yann LEROY, yann.leroy@centralesupelec.fr, Laboratoire Genie Industriel, CentraleSupélec, Université Paris-Saclay, Gif-sur-Yvette, France;
Emilie POIRSON, emilie.poirson@ec-nantes.fr, IRCCyN, Ecole Centrale de Nantes, Nantes, France

## Abstract

This research set out to identify and structure from online reviews the words and expressions related to customers' likes and dislikes to guide product development. Previous methods were mainly focused on product features. However, reviewers express their preference not only on product features. In this paper, based on an extensive literature review in design science, the authors propose a summarization model containing multiples aspects of user preference, such as product affordances, emotions, usage conditions. Meanwhile, the linguistic patterns describing these aspects of preference are discovered and drafted as annotation guidelines. A case study demonstrates that with the proposed model and the annotation guidelines, human annotators can structure the online reviews with high inter-agreement. As high inter-agreement human annotation results are essential for automatizing the online review summarization process with the natural language processing, this study provides materials for the future study of automatization.

## 1. Introduction

The development of e-commerce has generated a massive amount of online review data. Customers can talk about all aspects of a product in online reviews. These reviews are useful not only for potential buyers making purchase decisions, but also for designers collecting user requirements and preferences (Raghupathi, Yannou, Farel, & Poirson, 2015). Unlike focus group exercises or surveys that are based on physical prototypes, the large amounts of readily accessible review data enable designers to acquire the full spectrum of customer needs in a timely and efficient manner. Studies of the review data have accordingly been devoted to predicting trends, adjusting market strategy, improving the design of next-generation products, etc., to meet customer requirements more closely and enhance customer satisfaction.

Entering the data era, profiting from the big data is an essential ability for today's product designers, especially for those who must continually evolve their product in the competitive market. Nevertheless, the unstructured nature of the raw text data is a stumbling block to make good use of the information in the online reviews. It is difficult to identify and structure the words and expressions that are meaningful for product design in a text. Meanwhile, due to the large quantity of online review data, it is extremely time and resource consuming to process the data with only human effort. The computer is needed in online review analysis.

With the development of natural language processing techniques, various sentiment analysis methods have been applied to automatically summarize users' satisfaction and dissatisfaction (Ravi & Ravi, 2015). Most of these methods are feature-based: they extract the words and expressions related to the product features and the opinion words associated with these product features (Burnap et al., 2016; M. Hu & Liu, 2004; Tuarob & Tucker, 2015b; Zhang, Sekhari, Ouzrout, & Bouras, 2016). An implicit underlying assumption of these studies is that what customers like and dislike are the product features.

However, these studies have issues at the theoretic level. Reviewers express their like and dislike not only on product features. Also, they are focused on how the product performs when they use the product in a certain environment, whether it can help them achieve their goals, what their first impression of the product is. Answers to these questions are important for designers to better understand why the user like or dislike their product. For example, when customers require a larger screen for their cellphone, designers are more concerned with why customers need a larger screen. The feature-based sentiment analysis provides limited information. Meanwhile, the complexity of the linguistic patterns used in the text is not addressed. Two human annotators are highly likely to obtain different summarization results when annotating words and expressions related to user requirements. Yet high inter-agreement of annotation data is necessary for the study of automatic summarization with the computer. These issues must be addressed before processing an automatic online review analysis.

Therefore, in this paper, we first conducted an extensive literature review in the domain of design science. The previous studies in design science are summarized to find out the concepts that are widely used in product development. According to the literature review, during product design, designers analyze user preference and requirements from different angles. In fact, besides users' preference on product features, designers are also focused on users' requirement on product affordance, usage condition, user emotion and perception. In the second step, we manually process a sample of 265 review sentences to study whether the reviewers express their requirements on these concepts and to recognize what the linguistic patterns are when reviewers describe these concepts. Linguistic patterns are defined as how the concepts are described semantically and syntactically (Zouaq, Gasevic, & Hatala, 2012). These patterns are important as they can be used as heuristic rules in the future study of automatizing the online review summarization process. In the third step, to evaluate the linguistic patterns, we drafted them as annotation guidelines. Two human annotators are asked to manually process the sample of review data based on the annotation guidelines. Then, the inter-agreement among the human annotators' summarization results and our annotation results are calculated.

The main contributions of this study are: 1) To the best of our knowledge, we are the first to study user requirements on the entities that are expressed in online reviews other than the product feature. Comparing with the previous research in online review analysis, we extract more useful information for product development from online reviews. Our summarization model is constructed based on the design methods and design models proposed in the domain of design science, it therefore has a theoretical basis. Designers can directly apply the summarized information to the design methods and models that they use. 2) As foundation work, the linguistic patterns and the high inter-agreement human annotation results provided in this research are essential for automatizing the summarization process in future studies. The linguistic patterns, which are programmable with the help of natural language toolkits, can be used as heuristic rules for automatic summarization. The human annotation results can be used as ground truth. By comparing the ground truth and the automatic summarization results, the performance of the automatic summarization method can be evaluated.

In what follows, Section 2 reviews the previous studies in online review analysis. Section 3 reviews the research in design science and summarize the key concepts that are widely used in product development. Section 4 describes the research framework. Section 5 presents the proposed summarization model and a manual summarization of a sample of 256 online review sentences. Section 6 presents the linguistic patterns that are recognized based on the manual summarization. Section 7 describes the experiment to evaluate the linguistic patterns, and Section 8 presents a general conclusion.

## 2. Online review analysis for product design

Based on our extensive literature review, the current online review summarization models for design fall into two groups: feature-based summarization, and other-entity-based summarization (Table 1).

### 2.1 Feature-based summarization

Researchers in computer science M. Hu and Liu (2004) were the first to propose a feature-based summarization method with natural language processing. The objective was to assist a range of readers: potential buyers, manufacturers, designers, etc. Their method identifies feature chunks (i.e. words and expressions) associated with opinion. The feature is defined as a product feature, attribute or function. The opinion is defined as a users' subjective evaluation of the feature. A sentiment lexicon is then used to sign the opinion as positive or negative. Bagheri, Saraee, and de Jong (2013), Wang et al. (2014) and Kang and Zhou (2017) later iteratively improved the accuracy of the automatic summarization. These studies focused more narrowly on the performance of the computer algorithm; how to analyze such structured data to provide design-related insights was not discussed.

Unlike the researchers in computer science, the researchers in design science are more interested in what conclusions for product design can be drawn from the summarized data. (Chung & Tseng, 2012) used the association rule mining algorithm to learn the most satisfying product features to promote business intelligence. Tuarob and Tucker (2013, 2014, 2015b); Tucker and Kim (2011) proposed a series of methods to monitor consumer trends, predict product longevity, identify the lead user and quantify product favorability based on features and opinions mined from reviews. (Lee & Choeh, 2014) conducted a study to predict the usefulness of online reviews based on the summarized features and opinions. Suryadi and Kim (2016) summarized the features mentioned in reviews to analyze the relations between product features and sales rank. Zhang et al. (2016) proposed a method to devise a product improvement strategy by analyzing reviewers' sentiment polarity on different product features. Most recently, Jin, Ji, and Yan (2017) proposed a method to select representative reviews based on the identification of chunks relevant to product features. (Law, Gruss, & Abrahams, 2017) used the supervised machine learning algorithm to discover automatically the defects of dishwasher appliance based on product features and user opinion.

**2.2 Other-entity-based summarization**

Various applications of the feature-based summarization have been described. However, product features cannot cover all the significant issues addressed in customer reviews. Some critical information needed to understand the rationale of purchase and related decision-making processing is ignored in the feature-based analysis method. (Min & Park, 2012) proposed a method to mesure the helpfulness of online reviews by customer's mentions about experiences. In their method, customer experience was identified based on time expression and product feature. Tuarob and Tucker (2015a) found that in many scenarios, product features expressed by customers were implicit in nature and did not directly map to engineering design targets. They therefore proposed a method to translate sentences in social media text into a set of keywords representing customer preference, based on a co-word network generated from the review text. Jin, Ji, and Kwong (2016) found that sentiment analysis at the feature level failed to give enough information on reasons for dissatisfaction. They thus proposed a method to represent the detailed reasons stated in the reviews by a set of words. However, the co-word network was so large, and the concept of the detailed reason is so general that the resulting set of words was difficult to understand. For example, "I have to squint the screen to read this on Nokia N9" is translated to {screen, angle, iPhone, need fix upgrade, bigger, larger, extra}, which is not understandable as a customer's preference. Zhou, Jianxin Jiao, and Linsey (2015) proposed a method to identify use cases from online reviews to gain a better understanding of latent customer needs. The use case in their research is defined by three entities: user type, interaction environment and contextual events. However, their method requires large use case databases, which are not yet available.

Table 1. Summarization models proposed in the literature and their applications to design

| Model | Authors | Application |
|---|---|---|
| **Feature-based** | | |
| Product feature and opinion | Hu & Liu (2004) | |
| Product feature | Tucker & Kim (2011) | Design trends monitoring |
| Product feature | Chung & Tseng (2012) | Building market strategies |
| Product feature and opinion | Zhan et al. (2009) | Gathering customer concerns |
| Product feature and opinion | Tuarob & Tucker (2013) | Product longevity prediction |
| Product feature | Tuarob & Tucker (2014) | Lead user identification |
| Product feature | Lee, et al. (2014) | Predicting the helpfulness of online reviews |
| Product feature and opinion | Tuarob & Tucker (2015) | Product favorability quantification |
| Product feature | Suryadi & Kim (2016) | Sales rank prediction |
| Product feature and opinion | Zhang et al. (2016) | Construction of product improvement strategy |
| Product feature and opinion | Law et al. (2017) | Discovering product defects |
| **Other entities** | | |
| Customer experience | Min et al. (2012) | Predicting the helpfulness of online reviews |
| Customer preference | Tuarob & Tucker (2015) | Implicit customer preference identification |
| Use cases | Zhou et al. (2015) | Latent customer needs elicitation |
| Unsatisfied feature – reasons | Jin et al. (2016) | Dissatisfaction reason investigation |
| Informative words | Singh et al. (2017) | Online review disambiguation |
| Dimensions of quality | Tirunillai & Tellis (2014) | Dimensions of quality summarization |

| **Our proposed model** | |
|---|---|
| Feature | |
| Affordance | |
| Perception | Summarization of multiple aspects of user needs |
| Emotion | User requirements' correlation analysis |
| Usage condition | |

## 3. Key concepts in design models

Based on a thorough literature review, five key concepts were summarized from the current studies in design science. During product development, designers are focused on user preference and requirements related to these five concepts.

### 3.1 Product feature

Product development starts with the collection of user requirements (Eppinger & Ulrich, 2015). However, no matter what the requirements are, the design of the product eventually converges to the determination of product features, i.e. the components of the product and their physical characteristics. The words and expressions related to product features depend on the product. B. Liu (2012) built a two-level hierarchical representation model to conceptualize the relation between product features. "Two-level" refers to component and attribute levels. There is a "has property" relation between these two levels. For example, "screen" as a component of a smartphone has an attribute "resolution" as a property. "Hierarchical" refers to the different levels of decomposition of the product components. There is a "part of" relation between these levels. For example, "background light" is a part of the "screen", "screen" is a part of "smartphone".

### 3.2 Affordance

The affordance-based design is focused on the potential behavior between the user and the product. J. R. Maier and Fadel (2003) pointed out that during product development, designers should favor the beneficial affordances of the product, while reducing the harmful affordances of the product.

The concept of affordance was introduced into product design because of the deficiency of the product function model. The product function model has been widely used in product design ever since design science began. Lawrence D. Miles proposed the method of functional analysis as part of his method for value analysis in 1947. In much work, the function has been described as transforming material, energy or signal, or as an abstraction of behavior, or as a transformation of input into output (J. Maier & Fadel, 2001).

However, J. R. Maier and Fadel (2003) found that the function model is unsuited to the design of products other than the mechanical systems of a transforming character, as such products cannot be represented in an input/output model. For example, the design of a chair for sitting on does not involve any transformation. Also, a function-based approach is unsuited to products where humans are involved as active users, because functions model the workings of a product, not its interaction with people. While affordance, defined as a relationship between two subsystems in which potential behaviors can occur that would not be possible with either subsystem alone (J. R. Maier & Fadel, 2009), can tackle these issues.

Describing the difference between function and affordance has prompted much discussion (Brown & Blessing, 2005; Brown & Maier, 2015; Kannengiesser & Gero, 2012). Although still debated, the consensus is that affordances do not include the notion of teleology. More specifically, functions refer to what a product is designed to do, while affordances refer to what users do with the product. Therefore, the usages of the product described in online reviews should conceptualize as product affordances, as reviewers tell their real stories of using the product.

Hu and Fadel (2012) and Mata et al. (2015) summarized three affordance description forms to describe product affordances (Table 2): "verb-ability", "verb + noun-ability" or "verb (+ object)".

Table 2 Existing affordance description forms, summarized by (J. Hu & Fadel, 2012)

| Form | Alternative form | Example |
|---|---|---|
| Verb + -ability | | Grab-ability, wrist-ability |
| Verb + noun + -ability | Noun + verb + -ability | Lift handle-ability, rotate gear-ability |
| Transitive verb + noun | Intransitive verb | Collect water, lubricate part |

### 3.3 Emotion

Emotional design is focused on users' physiological needs on their emotion. The emotional design was first proposed by Norman (2004), who claimed that a designer should address the human cognitive ability to elicit appropriate emotions so as to obtain a positive experience. A positive experience may include positive emotions (e.g., pleasure, trust) or negative ones (e.g., fear, anxiety), depending on the context (for example, a horror-themed computer game).

Various emotional word lexicons were constructed in prior research (Bradley & Lang, 1999; Mohammad & Turney, 2013; Scherer, 2005; Strapparava & Valitutti, 2004). For example, Plutchik (1994) proposed eight primary emotions, grouped by positive-negative opposites: joy versus sadness; anger versus fear; trust versus disgust; and surprise versus anticipation. However, there is still no standard way to categorize emotions (Ekman (1992).

### 3.4 Perception

Semantic differential methods are focused on the users' physiological needs on the perception of the product. These methods are widely used when developing the form of the product (Hsu, Chuang, & Chang, 2000; Petiot & Yannou, 2004). It concerns the symbolic qualities of man-made forms in their context of usage. For example, a person may describe a glass with the words "modern", "traditional", "fragile", "strong", etc.

Hsu et al. (2000) and Petiot and Yannou (2004) collected 24 pairs of adjectives to describe users' perception of telephones and 17 pairs of adjectives on table glasses. It was found that perceptions were described with adjectives usually paired with antonyms.

### 3.5 Usage condition

Usage condition is also called usage condition or usage environment. It comprises all the factors characterizing an application and the environment in which a product is used (Green, Tan, Linsey, Seepersad, & Wood, 2005). Knowing usage conditions is important in design evaluation, usage scenario simulation, and user pain identification, because usage condition influences customer behavior through product performance, choice, and customer preference (Bekhradi, Yannou, Farel, Zimmer, & Chandra, 2015; Yannou, Cluzel, & Farel, 2016). Based on this observation, Yannou, Hen & He developed usage context-based design (He, Chen, Hoyle, & Yannou, 2012; He, Hoyle, Chen, Wang, & Yannou, 2010; Yannou et al., 2009; Yannou, Yvars, Hoyle, & Chen, 2013).

Various usage situation models have been proposed. Belk (1975) describes a model that splits user situations into five groups: task definition, physical surroundings, social surroundings, temporal perspective and user's antecedent states. Green et al. (2005) narrow down the scope of usage condition to two major components: application context and environment context. He et al. (2012) emphasized that usage condition covers all aspects related to the use of a product but excludes customer profile and product attributes.

### 3.6  Observation and discussion

First, each design model clearly takes a different angle in translating user requirements. Focusing solely on product features does not therefore enable designers to perform a comprehensive analysis of the weaknesses and strengths of their product.

Second, in previous studies of online review analysis, the concept of perception, emotion and satisfaction are confused. Based on our literature review, these three concepts are different at the psychological level. Perception represents how the customer thinks about the product. the human and the product are both involved in the generation of the perception. While emotion denotes the psychological state of the customer. It can arise solely or result from a perception or satisfaction. The polarity of the perception or the polarity of the emotion does not always match the polarity of satisfaction. For example, people may be satisfied with a product that makes them feel fear (e.g. thriller movies), although "fear" is generally considered as a negative emotion. Also, "low battery capacity" does not necessarily mean that the customer dislikes the battery, although "low" is generally considered as a negative perception. A customer who is used to carry a power bank can tolerate this feature.

Finally, comparing with function, affordance is better suited to online review summarization. Function reasoning works well when a device is unknown, i.e. assigned a function, to predict possible devices (Brown & Blessing, 2005). However, online reviews describe users' experience with existing products. How the customers use the product is not necessarily as the designers intended: knowledge of misuse and novel usages can help product improvement and innovation.

### 4.  Research framework

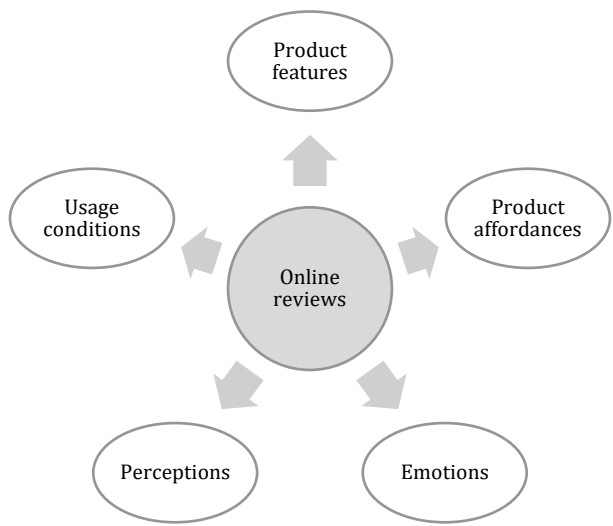

Figure 1. Our proposed summarization model

The proposed design-centered summarization model is based on the five key concepts we identified from the literature: product feature, perception, emotion, product affordance and usage condition (Figure 1).

The research framework is illustrated in Figure 2. To identify design-related concepts from online reviews, it is important to recognize the linguistic patterns when the reviewers describe these concepts in the review text. Indeed, meaningful words and expressions are not isolated in a review sentence. For example, Figure 3 shows the dependent relationship between feature and opinion. Thus, we must define the relationship between these concepts. The authors set out to manually label the words and expressions related to the five concepts in raw online review data corpus. Then, linguistic patterns are discovered by observing the summarization results. Next, the linguistic patterns are drafted as annotation guidelines. Two human annotators are asked to annotate the online review sample. Finally, the inter-agreement among the two human annotators and our annotation results are calculated to evaluate the linguistic patterns.

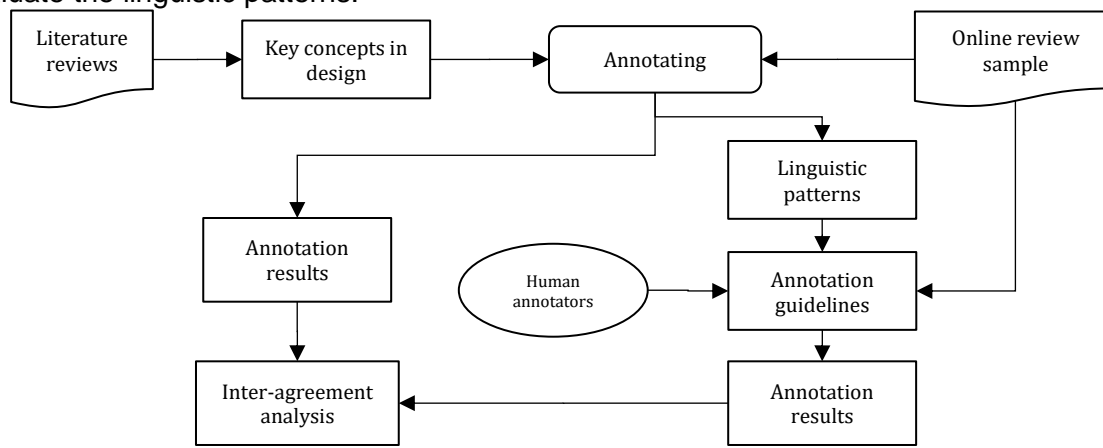

Figure 2. Research framework

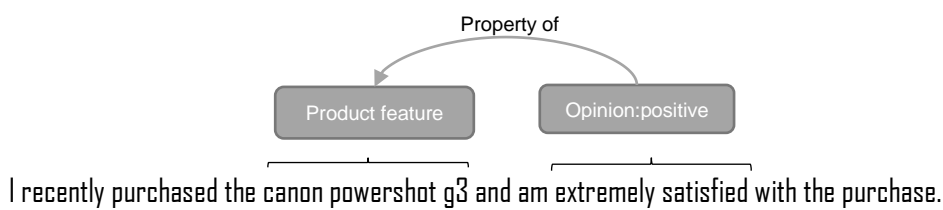

Figure 3. Example of annotation task

## 5. Manual summarization

## 5.1  Data preparation

265 review sentences of Kindle Paperwhite 3 e-reader (hereafter referred to as KP3) are downloaded from Amazon.com. These sentences come from the first 10 reviews of the KP3. All the 10 reviews were badged "verified purchase", which ensured their authenticity. The 265 sentences contain 4766 words in all. Table 3 gives detailed information for each review.

Table 3. Detailed information for each review

| Review number | Number of sentences | Number of words | Star rating | Date published | Number of "helpful" votes |
|---|---|---|---|---|---|
| 1 | 38 | 546 | 1 | Jul 21, 2015 | 852 |
| 2 | 9 | 147 | 1 | Jul 3, 2015 | 529 |
| 3 | 24 | 320 | 3 | Oct 12, 2015 | 144 |
| 4 | 36 | 684 | 5 | Jul 4, 2015 | 160 |
| 5 | 3 | 36 | 5 | Oct 17, 2015 | 78 |
| 6 | 51 | 909 | 5 | Jul 17, 2015 | 137 |
| 7 | 17 | 336 | 2 | Jul 24, 2015 | 94 |
| 8 | 29 | 684 | 1 | Jul 2, 2015 | 465 |
| 9 | 25 | 508 | 4 | Jul 8, 2015 | 154 |
| 10 | 33 | 596 | 5 | Aug 8, 2015 | 32 |

Before annotating, two basic rules were made for the sake of consistency: (i) articles "a(n)", "the" were not considered in the annotation, and (ii) pronouns such as "it", "them" were resolved and annotated when relevant to the concept.

## 5.2  A brief look at the summarized data

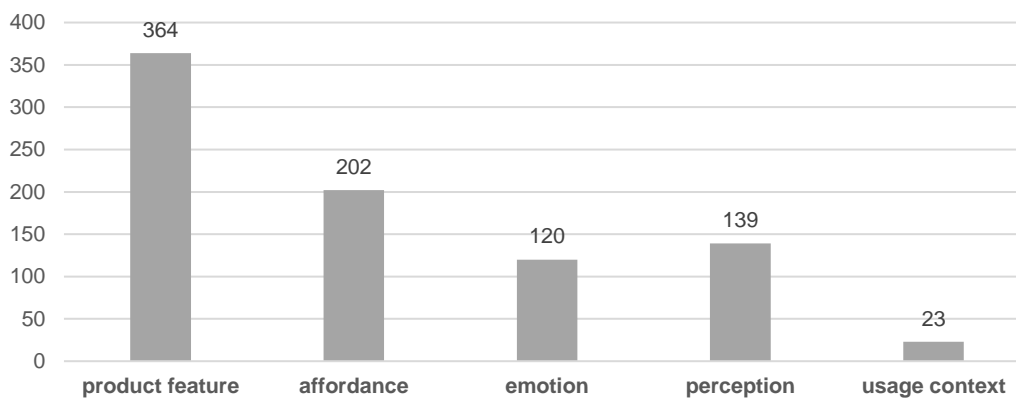

Figure 4. The descriptive statistics of the summarization results

It can be seen from the statistical data (Figure 4) that besides product feature, a large number of words and expressions are labeled in the review data sample, showing that our summarization model does provide designers more design-related information. Therefore, our proposed summarization modelallows designers to make better use of the information described in the online reviews to support product development.

Table 4. A sample of the summarization results

| Sentence | Structured data |
|---|---|
| However, as soon as I received it, I noticed a line of dead pixels right in the center of the screen (Note pic #1). | Product feature: {it, pixels, screen}, Affordance: {ability to receive it, ability to notice a line of dead pixels}, Perception: {dead (pixels)} |
| There's a significant amount of dust and unrecognizable particles under the screen. | Product feature: {significant amount of dust, unrecognizable particles, screen} Perception: {significant amount (dust), unrecognizable (particles)} |
| For those who hesitantly bought this device because of the boasted 300ppi screen and thought it would be on par with the Kindle Voyage, think again, it's not! | Product feature: {this device, 300 ppi screen, it, Kindle Voyage} Affordance: {ability to buy this device} Perception: {boasted (300 ppi)} |
| The setup is extremely easy. | Affordance: {ability to setup} Perception: {extremely easy (setup)} |
| I am so excited to be able to finally read ebooks in the sun outside and to read in bed at night without killing my eyes or keeping the husband up. | Affordance: {ability to read ebooks, ability for I to read, ability for killing eyes, ability for keeping up husband} Emotion: {excited} Perception: {not (kill, keep)} Usage condition: {in sun outside (read), in bed at night (read)} |

Table 4 shows the words and expressions summarized from five sentences. Multiple ways of visualizing the summarized data can be developed to gain insights for product design. For example, co-occurrence maps can be created to analyze the correlation among the extracted product features, product affordances, perceptions, emotions and usage conditions. In the map, the weight of the node represents the frequency of occurrence. The width of the edge represents the frequency of co-occurrence of two concepts at the sentence level.

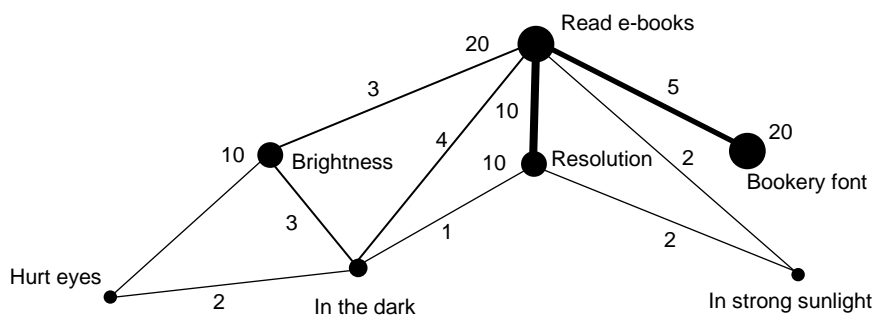

Figure 5. Correlation analysis of affordance, usage condition and product feature

### 5.3 Implications for designers

Figure 5 illustrates a part of the co-occurrence map. In the map, "read e-books" and "not hurt eyes" are affordances of the Kindle e-reader. "Brightness", "resolution" and "bookery font" are product features. "In the dark" and "in strong sunlight" are usage conditions". The affordance "read e-books" is determined by the three product features, as they are all connected with the affordance. The most influential product feature to the affordance "read e-books" is the resolution, as the segment between these two nodes is the thickest. Therefore, to improve the reading experience, designers have to consider these three product features, most importantly the resolution. However, to improve the reading experience in a dark environment, designers have to pay more attention to the brightness, as more reviewers mentioned the brightness when talking about using e-readers in the dark. Moreover, when adjusting the brightness feature, designers need to consider its influence on the affordance "hurt eyes", as "brightness" is also linked with this affordance.

In this way, it is easy for designers to learn the correlation between different aspects of user requirements, which supports decision making when designing new e-readers.

## 6. Linguistic patterns

This section describes the linguistic patterns that we find out by observing the manual summarization result. Here the linguistic pattern is defined as how the concepts are described syntactically and semantically (Zouaq et al., 2012). For example, for identifying the words and expressions related to product features, one of the linguistic patterns that are widely used is that product features are described by nouns having a high frequency. In this linguistic pattern, "noun" is regarded as a syntactical feature of the natural language.

### 6.1 Product feature

For product feature, two adjustments were made based on the two-level hierarchy model proposed by B. Liu (2012). First, the scope of component and attribute in the hierarchy model was enlarged. The words and expressions describing the things physically attached to the product (e.g. particles under the screen, cover of Kindle, e-books, Amazon account), or the things produced by the product (e.g. defects, issues), or the dimension of the attribute (e.g. difference in clarity, variation of color) were all labeled as product feature. These chunks appeared frequently in the reviews and would help designers understand the summarized results. Second, whereas most research (M. Hu & Liu, 2004; B. Liu, 2012; Zhang et al., 2016) has only considered noun chunks as relevant to product features, in our summarization model, linking verbs were also taken into account. For example, in the sentence "This NEW Kindle looked great", "look" is labeled product feature as it referred to the appearance of the Kindle.

Therefore, the linguistic pattern for identifying product features are:

1) the nouns that describe product component and attribute;
2) the linking verbs adjacent to a product name or product component.

## 6.2 Affordance

Hu and Fadel (2012) summarized from the literature that the affordances can be described in three forms: "verb-ability", "verb + noun-ability", "verb (+ noun)". For example, a chair affords "sit-ability", an e-reader affords "read book-ability", a pen affords "writing". It can be seen that the verb is an indispensable element in the affordance description. However, first, we find that in the online reviews, nouns and adjectives can also describe affordances, especially nouns and adjectives that are derived from verbs, having the suffix "-able", "-ible", "-ity(-itiies)". For example, "movability" of a chair, "transportability" of an e-reader. Second, not all verbs are product affordances, especially emotional verbs and stative verbs. Instead of a potential behavior between the user and the product, they describe solely the psychological state of the reviewer and the state of the product. For example, in the sentence "It looks nice", the word "looks" only describes the appearance of the product. In the sentence "I want to have the e-reader", the word "want" only describes the cognition of the reviewer. Third, we find that in the online reviews, reviewers talk not only about the product, but also about logistics and after-sales service. These words are not affordances of the product, as the product is not involved in the action. For example, in the expression "I contact the after sales team", the word "contact" is not labeled as affordance.

Therefore, we use the description form "ability to [action word] [action receiver]" to structure the affordances described in online reviews. The linguistic patterns for identifying product affordances are:

1) The verbs are action words, except stative verbs, emotional verbs and the verbs describing an action in which the product is not involved;
2) The nouns and adjectives, derived from verbs, having the suffix "-able", "-ible", "-ity(-ities)", are action words
3) Action receiver is the object of the action word

## 6.3 Emotion

As discussed in Section 3.3, various emotional lexicons were constructed in prior research (Bradley & Lang, 1999; Mohammad & Turney, 2013; Scherer, 2005; Strapparava & Valitutti, 2004). Therefore, identifying emotional word is relatively straightforward, as these lexicons can be directly used. We find that first, emotional words are not only adjectives. They can also be verbs and nouns. For example, in the sentence "I hope to have an e-reader for a long time", the word "hope" denoted the emotional state of the reviewer, i.e., desire. Second, as the emotional word describes the emotional state of human, the emotional word should be adjacent to the words describing humans. For example, in the sentence "The color of the chair is exciting", the word "exciting" is not an emotional word, as it describes the property of the chair, i.e., color. However, in the sentence "The color of the chair makes me excited", the word "excited" is labeled as a emotional word.

Therefore, the linguistic pattern for identifying emotions is:
1) Words in emotional lexicons, adjacent to the words describing humans.

## 6.4 Perception

Perception is defined as the way in which the product is regarded, understood, or interpreted by the reviewer. It means that when the reviewer describes their perception, there must be at least one object that receives the perception. Meanwhile, as summarized in Section 3.2, perceptual words are adjectives paired with antonyms. Therefore, perceptual words are the adjectives adjacent to product features or the adverbs adjacent to product affordances, having antonyms. For example, in the expression "short battery life", the word "short" is the perception of the product feature "battery life". In the sentence "I can read the book easily", the word "easily" is annotated as the perception of the action word "read". In addition, perceptual words can be a negation. For instance, in the sentence, "I cannot listen to music", "cannot" is a perceptual word, meaning that the product does not have the ability to for the user to listen. However, not all adjectives are perceived configurations, especially those adjectives in proper nouns. For example, the word "internal" in "internal storage" does not describe a perception.

Therefore, the linguistic pattern for identifying perceptions is:

1) Adjectives adjacent to product features, or adverbs adjacent to product affordances, having antonyms, except the adjectives and adverbs in a proper noun.

**6.5 Usage condition**

Usage condition is defined as all the factors characterizing an application and the environment in which a product is used. Consequently, the words and expressions describing usage conditions are adjacent to product affordance.

Based on our observation, reviewers mainly talk about physical surroundings when they use the product. Therefore, the words and expressions describing usage conditions usually begin with the preposition of place, such as "on", "above", "in", "at". For example, "read book at night", "read book in bed". Therefore, the linguistic pattern for identifying usage conditions is:

1) Prepositional phrases adjacent to product affordances, having preposition of place.

## 7. Evaluating the linguistic patterns

### 7.1 Data preparation and participants

We drafted annotation guidelines based on the linguistic patterns that we discovered from the manual summarization results. The guidelines contain linguistic patterns and examples to explain the annotation task. A Q&A section helps annotators quickly locate the answer to questions they may have during the annotation.

To evaluate the linguistic patterns, two Ph.D. students in design science were asked to annotate carefully the 265 online KP3 reviews independently (see Section 3.2 for the data preparation) following the guidelines we drew up. After finishing the independent annotation, the two annotators compared their results and discussed the differences in their annotation results. If a difference was due to an error made by one annotator, then the students were asked to correct the result.

### 7.2 Evaluation metrics

The quality of the linguistic patterns was evaluated by the inter-agreement (Pustejovsky & Stubbs, 2012) of the two student annotators' results and the authors' results. The inter-agreement denotes how often the annotators agree with each other. Obviously, high inter-agreement means that the annotators' results were precise and accurate in comparison with the authors' results, and thus signifies that the linguistic patterns were well established, and the annotation guidelines were clearly drafted. Fleiss's kappa (Fleiss, 1971) is widely used to calculate the inter-agreement. The equation is:

$$K = \frac{Pr(a) - Pr(e)}{1 - Pr(e)},$$

where $Pr(a)$ is the relative observed agreement between annotators, and $Pr(e)$ is the expected agreement between annotators if each annotator was to randomly pick a category for each annotation. To interpret Fleiss's kappa, the scale proposed by Landis and Koch (1977) is used (Table 5).

Table 5. Interpreting Fleiss's kappa as proposed by Landis and Koch (1977)

| K | Agreement level |
|---|---|
| <0 | Poor |
| 0.01 – 0.20 | Slight |
| 0.21 – 0.40 | Fair |
| 0.41 – 0.60 | Moderate |
| 0.61 – 0.80 | Substantial |
| 0.81 – 1.00 | Perfect |

### 7.3 Results and discussion

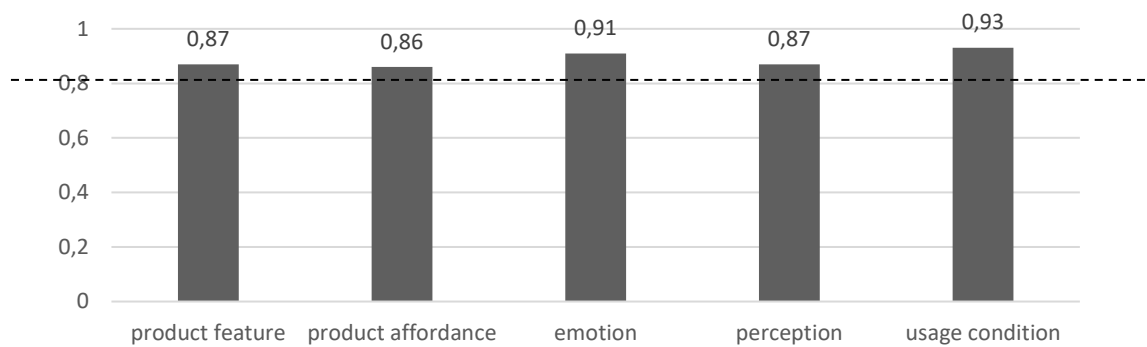

Figure 6. Fleiss's kappa for each concept

Figure 6 shows the Fleiss's kappa for each concept. It can be seen that the inter-agreement for all the concepts exceeded 0.8, which means that our guidelines were "perfect" on the scale of Landis and Koch (1977).

The differences between the two annotators' results and the researcher's result were analyzed, and three reasons were found. First, some sentences were unclear owing to the indeterminacy of natural language. This often occurred when a reviewer expressed a perception in the interrogative form. For example, in the sentence, "Second: The 300dpi thing is quite meh (in comparison to 212 and even 167 of the pw1), I mean, is it better? Will it make much of a difference?", it is difficult to tell whether the reviewer thinks the resolution is better or not. Second, the annotators reported misspelling as one reason for disagreement in the annotation. These two disagreements cannot be eliminated by improving the guidelines, as the problem is inherent in the review sentence. It is better to filter out these sentences before processing review summarization. Finally, annotators have different understandings of the concepts based on their knowledge. For example, one of the annotators considered that the word "setup" in the sentence "the setup is easy" was a product feature because the level of difficulty of the setup is an attribute of the software, while the other annotator regarded it as an action word that describes an affordance. Another example concerns the expression "be used to": whether it is an emotional word is still under discussion. These disagreements stem from the unclear definition of the concepts in design. With the development and clarification of the design models, they can be eliminated by fully listing the commonly-agreed lexicon related to each concept in the annotation guidelines (e.g. a database of affordances for each product).

## 8. Conclusion

### 8.1 Practical implications

In this study, we reviewed the current online review summarization methods for product design. Three main limitations were identified. First, extracting only features associated with opinions provides limited information for designers to understand user requirements. Today's designers are more closely focused on the interaction between product and user. Knowing how customers use a product, what their emotional state is when using it, and in what conditions they use it, is most important for best meeting customers' needs. Second, although other methods have been proposed to identify the words and expressions related to entities like customer concerns, novel usages, engineering characteristics, etc., the lack of common definitions in the domain of engineering design makes the meaning of summarized data obscure or ambiguous. Finally, the issue of complexity of linguistic patterns is not addressed, preventing the construction of high-quality manual summarization results as ground truth data in the future study of automatic online review summarization.

To overcome the limitations, we propose a multi-aspect online review summarization model based on an extensive literature review of the current design models. This model helps to manage the knowledge on user requirements. Five key concepts are identified: product feature, emotion and perception, affordance, usage condition. With a manual analysis of 265 review sentences of the Kindle Paperwhite 3 e-reader, the linguistic patterns describing these concepts are discovered. An experiment shows that linguistic patterns are well established. Comparing with previous research in online review analysis, with our proposed summarization model, designers can obtain more useful information from online review data, and thus make better use of the online reviews to help product development in data era.

Meanwhile, in our case study, we downloaded 265 review sentences regarding the Kindle e-reader from amazon.com. We manually annotated the words and expressions according to the proposed summarization template. We constructed a co-occurrence map to analyze the relations between different aspects of user needs. The map provides important information to support decision making during the development of new products. Designers are able to be aware of the influence of adjusting one product feature to other aspects of user requirement.

**8.2 Theoretical implications**

Today, people talk about text mining (Wamba, Akter et al. 2015). However, what information can be extracted from text data? To answer this question, we must use the correct domain theory to change the unstructured text data to structured data before further analysis. Sentiment analysis on product feature dominates the previous online review analysis for product design. However, both product feature and sentiment orientation lack a theoretical basis in design engineering. As previous research found, product features alone cannot cover all the significant issues addressed in customer reviews for product design. That is why we begin our study by reviewing the studies in the domain of design science.

**8.3 Limitations and perspectives**

A limitation of this research is that we analyzed only 265 reviews of one product in the case study. The small number of reviews thus caveats the conclusions. For perspectives, it is impossible to manually process all the online reviews one by one, as the quantity is huge. Therefore, the summarization process needs to be automated. Our research provides the essential materials for the study of automatization the online review summarization process, i.e., linguistic patterns and ground truth results, which can be directly used in the future study.

Acknowledgments: This research was supported by the China Scholarship Council, an organization of research scholarship.

Declarations of interest: none

## References

- Bagheri, A., Saraee, M., & de Jong, F. (2013). Care more about customers: Unsupervised domain-independent aspect detection for sentiment analysis of customer reviews. Knowledge-Based Systems, 52, 201-213. doi:10.1016/j.knosys.2013.08.011
- Bekhradi, A., Yannou, B., Farel, R., Zimmer, B., & Chandra, J. (2015). Usefulness Simulation of Design Concepts. Journal of Mechanical Design, 137(7), 071412. doi:10.1115/1.4030180
- Belk, R. W. (1975). Situational variables and consumer behavior. Journal of Consumer research, 2(3), 157-164.
- Bradley, M. M., & Lang, P. J. (1999). Affective norms for English words (ANEW): Instruction manual and affective ratings. Retrieved from
- Brown, D. C., & Blessing, L. (2005). The relationship between function and affordance. Paper presented at the ASME 2005 International Design Engineering Technical Conferences and Computers and Information in Engineering Conference.
- Brown, D. C., & Maier, J. R. (2015). Affordances in design. Artificial Intelligence for Engineering Design, Analysis and Manufacturing, 29(03), 231-234.
- Burnap, A., Pan, Y., Liu, Y., Ren, Y., Lee, H., Gonzalez, R., & Papalambros, P. Y. (2016). Improving Design Preference Prediction Accuracy Using Feature Learning. Journal of Mechanical Design, 138(7), 071404.

- Chen, L., Qi, L., & Wang, F. (2012). Comparison of feature-level learning methods for mining online consumer reviews. Expert Systems with Applications, 39(10), 9588-9601. doi:10.1016/j.eswa.2012.02.158
- Chung, W., & Tseng, T.-L. B. (2012). Discovering business intelligence from online product reviews: A rule-induction framework. Expert Systems with Applications, 39(15), 11870-11879.
- Ekman, P. (1992). An argument for basic emotions. Cognition & emotion, 6(3-4), 169-200.
- Eppinger, S., & Ulrich, K. (2015). Product design and development: McGraw-Hill Higher Education.
- Fleiss, J. L. (1971). Measuring nominal scale agreement among many raters. Psychological bulletin, 76(5), 378.
- Green, M. G., Tan, J., Linsey, J. S., Seepersad, C. C., & Wood, K. L. (2005). Effects of product usage context on consumer product preferences. Paper presented at the ASME Design Theory and Methodology Conference.
- He, L., Chen, W., Hoyle, C., & Yannou, B. (2012). Choice modeling for usage context-based design. Journal of Mechanical Design, 134(3), 031007.
- He, L., Hoyle, C., Chen, W., Wang, J., & Yannou, B. (2010). A framework for choice modeling in usage context-based design. Paper presented at the ASME 2010 International Design Engineering Technical Conferences and Computers and Information in Engineering Conference.
- Hsu, S. H., Chuang, M. C., & Chang, C. C. (2000). A semantic differential study of designers' and users' product form perception. International Journal of Industrial Ergonomics, 25(4), 375-391.
- Hu, J., & Fadel, G. M. (2012). Categorizing affordances for product design. Paper presented at the ASME 2012 International Design Engineering Technical Conferences and Computers and Information in Engineering Conference.
- Hu, M., & Liu, B. (2004). Mining and summarizing customer reviews. Paper presented at the Proceedings of the tenth ACM SIGKDD international conference on Knowledge discovery and data mining.
- Jin, J., Ji, P., & Gu, R. (2016). Identifying comparative customer requirements from product online reviews for competitor analysis. Engineering Applications of Artificial Intelligence, 49, 61-73. doi:10.1016/j.engappai.2015.12.005
- Jin, J., Ji, P., & Kwong, C. K. (2016). What makes consumers unsatisfied with your products: Review analysis at a fine-grained level. Engineering Applications of Artificial Intelligence, 47, 38-48. doi:10.1016/j.engappai.2015.05.006
- Jin, J., Ji, P., & Yan, S. (2017). Comparison of series products from customer online concerns for competitive intelligence. Journal of Ambient Intelligence and Humanized Computing, 1-16.
- Kang, Y., & Zhou, L. (2017). RubE: Rule-based methods for extracting product features from online consumer reviews. Information & Management, 54(2), 166-176.
- Kannengiesser, U., & Gero, J. S. (2012). A process framework of affordances in design. Design Issues, 28(1), 50-62.
- Landis, J. R., & Koch, G. G. (1977). The measurement of observer agreement for categorical data. biometrics, 159-174.
- Law, D., Gruss, R., & Abrahams, A. S. (2017). Automated defect discovery for dishwasher appliances from online consumer reviews. Expert Systems with Applications, 67, 84-94.
- Lee, S., & Choeh, J. Y. (2014). Predicting the helpfulness of online reviews using multilayer perceptron neural networks. Expert Systems with Applications, 41(6), 3041-3046. doi:10.1016/j.eswa.2013.10.034
- Liu, B. (2012). Sentiment analysis and opinion mining. Synthesis lectures on human language technologies, 5(1), 1-167.
- Liu, Y., Jin, J., Ji, P., Harding, J. A., & Fung, R. Y. K. (2013). Identifying helpful online reviews: A product designer's perspective. Computer-Aided Design, 45(2), 180-194. doi:10.1016/j.cad.2012.07.008
- Maier, J., & Fadel, G. (2001). Affordance: The Fundamental Concept in Engineering Design, ASME DETC/DTM, Pittsburgh, PA, Paper No. Retrieved from

- Maier, J. R., & Fadel, G. M. (2003). Affordance-based methods for design. Paper presented at the ASME 2003 International Design Engineering Technical Conferences and Computers and Information in Engineering Conference.
- Maier, J. R., & Fadel, G. M. (2009). Affordance based design: a relational theory for design. Research in Engineering Design, 20(1), 13-27.
- Min, H.-J., & Park, J. C. (2012). Identifying helpful reviews based on customer's mentions about experiences. Expert Systems with Applications, 39(15), 11830-11838. doi:10.1016/j.eswa.2012.01.116
- Mohammad, S. M., & Turney, P. D. (2013). Crowdsourcing a word–emotion association lexicon. Computational Intelligence, 29(3), 436-465.
- Norman, D. A. (2004). Emotional design: Why we love (or hate) everyday things: Basic Civitas Books.
- Petiot, J.-F., & Yannou, B. (2004). Measuring consumer perceptions for a better comprehension, specification and assessment of product semantics. International Journal of Industrial Ergonomics, 33(6), 507-525. doi:10.1016/j.ergon.2003.12.004
- Plutchik, R. (1994). The psychology and biology of emotion: New York, NY, US: HarperCollins College Publishers.
- Pustejovsky, J., & Stubbs, A. (2012). Natural Language Annotation for Machine Learning: A guide to corpus-building for applications: " O'Reilly Media, Inc.".
- Raghupathi, D., Yannou, B., Farel, R., & Poirson, E. (2015). Learning from product users, a sentiment rating algorithm Design Computing and Cognition'14 (pp. 475-491): Springer.
- Ravi, K., & Ravi, V. (2015). A survey on opinion mining and sentiment analysis: Tasks, approaches and applications. Knowledge-Based Systems, 89, 14-46. doi:10.1016/j.knosys.2015.06.015
- Scherer, K. R. (2005). What are emotions? And how can they be measured? Social science information, 44(4), 695-729.
- Strapparava, C., & Valitutti, A. (2004). Wordnet affect: an affective extension of wordnet. Paper presented at the Lrec.
- Suryadi, D., & Kim, H. (2016). Identifying the Relations Between Product Features and Sales Rank From Online Reviews. Paper presented at the ASME 2016 International Design Engineering Technical Conferences and Computers and Information in Engineering Conference.
- Tuarob, S., & Tucker, C. S. (2013). Fad or here to stay: Predicting product market adoption and longevity using large scale, social media data. Paper presented at the ASME 2013 International Design Engineering Technical Conferences and Computers and Information in Engineering Conference.
- Tuarob, S., & Tucker, C. S. (2014). Discovering next generation product innovations by identifying lead user preferences expressed through large scale social media data. Paper presented at the ASME 2014 International Design Engineering Technical Conferences and Computers and Information in Engineering Conference.
- Tuarob, S., & Tucker, C. S. (2015a). A product feature inference model for mining implicit customer preferences within large scale social media networks. Paper presented at the ASME 2015 International Design Engineering Technical Conferences and Computers and Information in Engineering Conference.
- Tuarob, S., & Tucker, C. S. (2015b). Quantifying product favorability and extracting notable product features using large scale social media data. Journal of Computing and Information Science in Engineering, 15(3), 031003.
- Tucker, C. S., & Kim, H. M. (2011). Trend mining for predictive product design. Journal of Mechanical Design, 133(11), 111008.
- Wang, T., Cai, Y., Leung, H.-f., Lau, R. Y. K., Li, Q., & Min, H. (2014). Product aspect extraction supervised with online domain knowledge. Knowledge-Based Systems, 71, 86-100. doi:10.1016/j.knosys.2014.05.018
- Xu, K., Liao, S. S., Li, J., & Song, Y. (2011). Mining comparative opinions from customer reviews for Competitive Intelligence. Decision Support Systems, 50(4), 743-754. doi:10.1016/j.dss.2010.08.021

- Yannou, B., Cluzel, F., & Farel, R. (2016). Capturing the relevant problems leading to pain and usage driven innovations: the DSM Value Bucket algorithm. Concurrent Engineering: Research and Applications, 1-16.
- Yannou, B., Wang, J., Rianantsoa, N., Hoyle, C., Drayer, M., Chen, W., . . . Mathieu, J.-P. (2009). Usage coverage model for choice modeling: principles. Paper presented at the ASME 2009 International Design Engineering Technical Conferences and Computers and Information in Engineering Conference.
- Yannou, B., Yvars, P.-A., Hoyle, C., & Chen, W. (2013). Set-based design by simulation of usage scenario coverage. Journal of Engineering design, 24(8), 575-603.
- Zhang, H., Sekhari, A., Ouzrout, Y., & Bouras, A. (2016). Jointly identifying opinion mining elements and fuzzy measurement of opinion intensity to analyze product features. Engineering Applications of Artificial Intelligence, 47, 122-139. doi:10.1016/j.engappai.2015.06.007
- Zhou, F., Jianxin Jiao, R., & Linsey, J. S. (2015). Latent Customer Needs Elicitation by Use Case Analogical Reasoning From Sentiment Analysis of Online Product Reviews. Journal of Mechanical Design, 137(7), 071401. doi:10.1115/1.4030159
- Zouaq, A., Gasevic, D., & Hatala, M. (2012). Linguistic patterns for information extraction in ontocmaps. Paper presented at the Proceedings of the 3rd International Conference on Ontology Patterns-Volume 929.